\documentclass{article}

\usepackage{PRIMEarxiv}

\usepackage[utf8]{inputenc} % allow utf-8 input
\usepackage[T1]{fontenc}    % use 8-bit T1 fonts
\usepackage{hyperref}       % hyperlinks
\usepackage{url}            % simple URL typesetting
\usepackage{booktabs}       % professional-quality tables
\usepackage{nicefrac}       % compact symbols for 1/2, etc.
\usepackage{microtype}      % microtypography
\usepackage{lipsum}
\usepackage{graphics,color}
\usepackage[pdftex]{graphicx}
\usepackage{latexsym,amsfonts,amsmath,tabularx,epsfig,enumerate,epstopdf,algorithm,algorithmic,empheq,subfigure,diagbox,caption}
\graphicspath{{media/}}     % organize your images and other figures under media/ folder

\def\Om{\Omega}
\def\Ga{\Gamma}
\def\pa{\partial}
\def\R{\mathbb{R}}
\def\S{\mathbb{S}}
\def\ol{\overline}

\def\dim{\mbox{dim}}

\def\bfx{\mbox{\boldmath$x$}}
\def\bfy{\mbox{\boldmath$y$}}
\def\bfn{\mbox{\boldmath$n$}}

\def\calN{\mathcal{N}}
\def\calB{\mathcal{B}}
\newtheorem{remark}{Remark}[section]
\newtheorem{theorem}{Theorem}[section]
\title{Learning Only On Boundaries: a Physics-Informed Neural operator for Solving Parametric Partial Differential Equations in Complex Geometries
}

\author{
  Zhiwei Fang \\
  \texttt{pentilm@outlook.com} \\
   \And
   Sifan Wang \\
  Graduate Group in Applied Mathematics \\
  and Computational Science \\
  University of Pennsylvania\\
  Philadelphia, PA 19104 \\
  \texttt{sifanw@sas.upenn.edu} \\
  \And
  Paris Perdikaris \\
  Department of Mechanichal Engineering \\
  and Applied Mechanics\\
  University of Pennsylvania\\
  Philadelphia, PA 19104 \\
  \texttt{pgp@seas.upenn.edu} \\
}

\begin{document}
\maketitle

\begin{abstract}
Recently deep learning surrogates and neural operators have shown promise in solving partial differential equations (PDEs). However, they often require a large amount of training data and are limited to  bounded domains. In this work, we present a novel physics-informed neural operator method to solve parametrized boundary value problems without labeled data. By reformulating the PDEs into boundary integral equations (BIEs), we can train the operator network solely on the boundary of the domain. This approach reduces the number of required sample points from $O(N^d)$ to $O(N^{d-1})$, where $d$ is the domain's dimension, leading to a significant acceleration of the training process. Additionally, our method can handle unbounded problems, which are unattainable for existing physics-informed neural networks (PINNs) and neural operators. Our numerical experiments show the effectiveness of parametrized complex geometries and unbounded problems.
\end{abstract}

% keywords can be removed

\section{Introduction}
The solution of partial differential equations is a crucial task in various disciplines, including physics, engineering, and many others. Although traditional numerical methods, such as finite element and finite difference methods, are widely used to solve PDEs \cite{oden2012introduction, brenner2008mathematical, hesthaven2007nodal, dolejvsi2015discontinuous, di2011mathematical}, they can be computationally demanding and resource-intensive for certain types of problems, such as those that are parameterized or unbounded.

Recently, machine learning methods such as physics-informed neural networks (PINNs) \cite{pinn, karniadakis2021physics, jagtap2021extended, meng2020ppinn, zhu2019physics, sanchez2020learning, tartakovsky2020physics, kissas2020machine, hennigh2021nvidia}, have emerged as a flexible way to solving PDEs and data assimilation. PINNs leverage neural networks to approximate the solution of a PDE and train the network by minimizing the residual of the PDE. These methods have demonstrated remarkable results in solving a wide range of PDEs, including nonlinear and high-dimensional problems. For example, Fang and Zhan \cite{fang2019deep} used PINNs to design electromagnetic meta-materials for user-specified targets and proposed a piece-wise design that can be applied in manufacturing. Raissi {\em et. al.} \cite{raissi2020hidden} developed hidden fluid mechanics to solve various physical and biomedical problems by extracting quantitative information that may not be directly measurable. Sun {\em et. al.} \cite{sun2020surrogate} employed a structured deep neural network to solve Navier–Stokes equations with applications in cardiovascular flows. Costabal {\em et. al.} \cite{sahli2020physics} applied active learning in PINNs and achieved lower error levels than random allocation. Kissas {\em et. al.} \cite{kissas2020machine} developed an application of PINNs for predicting arterial blood pressure from non-invasive 4D flow MRI.

While PINNs can be good predictors for PDEs' solutions, they cannot directly generalize to new scenarios for e.g. corresponding to new boundary conditions, geometries, etc.. This has motivated the development of neural operator architectures which leverage specialized neural network architectures to directly parametrize the solution operator of a PDE, instead of a single solution function.  Chen {\em et. al.} \cite{chen1995universal} showed a universal approximation theorem for operator and operator network architectures, which is one of the first works in operator learning. This framework was recently revived by Lu {\em et. al.} \cite{lu2021learning} who proposed the DeepONet architecture. In a parallel line of work, Li {\em et. al.} \cite{li2020fourier} drew motivation from the composition of linear and nonlinear layers in neural networks to propose a new class of neural operator architectures that also enjoy universal approximation guarantees. Other recent architectures include approaches based on PCA-based representations \cite{bhattacharya2020model}, random feature approaches \cite{nelsen2021random}, wavelet approximations to integral transforms \cite{gupta2021multiwavelet}, and attention-based architectures \cite{kissas2022learning}. 

Despite their widespread use, PINNs models are generally known for being hard to train and may give rise to several practical pathologies. Krishnapriyan {\em et. al.} \cite{krishnapriyan2021characterizing} elucidate a variety of potential failure scenarios pertinent to PINNs, providing a collection of examples that are highly relevant to both numerical researchers and engineers. Further, in their seminal work, Colton {\em et. al.} \cite{colton1998inverse} investigate challenges that arise in the numerical solution of acoustic wave scattering problems, which are widely used in practical contexts. Such scattering problems are typically posed in unbounded domains, rendering most existing PINNs formulations infeasible, due to their need to sample collocation points in an infinite domain. These challenges present a compelling motivation for the exploration of new methodologies that enable the application of PINNs to PDEs defined in unbounded domains.

In parallel to machine learning methods, boundary integral equations (BIEs) are a well-established method for solving PDEs, particularly for problems with complex geometries or unbounded domains. BIEs represent the solution of a PDE in terms of its boundary values, and can be solved using various techniques, such as the method of moments, collocation, and the Nystr{\"o}m method \cite{colton1998inverse}. This approach, which only requires boundary data, is particularly efficient in addressing infinite and semi-infinite problems in fields such as geomechanics, environmental science, physics, and engineering. As stated in \cite{sauter2010boundary}, BIEs provide a way to solve PDEs with unknowns only located on the boundary, and generate solutions anywhere in the domain with high accuracy. One of the numerical techniques for solving BIEs is boundary element methods (BEMs), which discretize the boundary of the domain and approximate the unknown function, referred to as potentials in many literature, on the boundary using finite elements. Many references, such as \cite{sauter2010boundary,katsikadelis2002boundary}, provide systematic introductions to BIEs and BEMs. Aussal {\em et. al.} \cite{aussal2022computing} studied the computation of singular integrals that appear in BEMs, illustrating with a scattering example. Colton {\em et. al.} \cite{colton1998inverse} reviewed the application of BEMs in inverse acoustic and electromagnetic scattering problems. Kagami {\em et. al.} \cite{kagami1984application} calculated the electromagnetic field using BEMs without any absorbing boundary conditions.

In this paper, we propose a novel machine learning-based solver that combines the strengths of BIEs and operator learning to address the challenges of solving PDEs with parameterized geometries and unbounded domains. By reformulating PDEs into BIEs, our method employs neural operator architectures to determine unknown potentials on the parameterized geometries using boundary conditions, and subsequently generates solutions anywhere in the domain with high accuracy. This approach reduces the training workload and is particularly efficient for solving problems with complex geometries. Additionally, by generating solutions through boundary information, it allows for the solutions of unbounded domain problems, which are not possible with traditional PINNs. We showcase the effectiveness of our method by applying it to various example problems and comparing the results with traditional solvers.

Our key contributions are summarized as follows.
\begin{enumerate}
    \item Boundary operator learner can avoid huge training points for large domains. This reduces time and spatial complexity for training and makes the proposed method's efficiency comparable to numerical solvers.
    \item Complex geometry is easy to handle since only boundary information is needed.
    \item Unbounded problem solver allows us to solve unbounded problems like bounded problems in terms of formulation and training cost, which cannot be solved by PINNs.
    \item BIE formulation reduces the order of derivatives in the PDEs, which simplifies the computational graph.
\end{enumerate}
The rest of this paper is organized as follows. In section \ref{sec:preliminaries}, we introduce some preliminaries that are related to the proposed algorithm. In section \ref{sec:BINOMAD}, we propose our algorithm in detail by using a motive example. Numerical experiments are shown in section \ref{sec:numerical} to verify our algorithm. Advantages, disadvantages, and future works are discussed in section \ref{sec:conclusion}.

\section{Preliminaries}\label{sec:preliminaries}
In this section, we recap some related topics in preparation for the proposed algorithm. We encourage readers to read the original papers cited below to get further details on these topics.

\subsection{Physics informed neural networks}
PINNs are a method for inferring a continuous latent function $u(\bfx)$ that serves as the solution to a nonlinear PDE of the form:
\begin{align}
    \calN[u](\bfx)    &=0\quad\mbox{ in }\Om,\label{eqn:pde1}\\
    \calB[u](\bfx)    &=0\quad\mbox{ on }\pa\Om,\label{eqn:pde2}
\end{align}
where $ \Om $ is an open, bounded set in $\mathbb{R}^d$ with a piecewise smooth boundary $\pa\Om$, $\bfx\in\mathbb{R}^d$, and $\calN$ and $\calB$ are nonlinear differential and boundary condition operators, respectively.

The solution to the PDE is approximated by a deep neural network, $u_\theta$, which is parameterized by $\theta$. The loss function for the network is defined as:
\begin{equation}
    L(u;\theta) = \frac{\omega_e}{N_p}\sum_{i=1}^{N_p}|\calN[u_\theta](\bfx^p_i)|^2 + \frac{\omega_b}{N_b}\sum_{i=1}^{N_b}|\calB[u_\theta](\bfx^b_i)|^2,\label{loss}
\end{equation}
where $ \{\bfx^p_i\}_{i=1}^{N_p} $ and $ \{\bfx^b_i\}_{i=1}^{N_b} $ are the sets of points for the PDE residual and boundary residual, respectively, and $\omega_e$ and $\omega_b$ are the weights for the PDE residual loss and boundary loss, respectively. The neural network $u_\theta$ takes the coordinate $\bfx$ as input and outputs the corresponding solution value at that location. The partial derivatives of the $ u_\theta $ with respect to the coordinates at $ \calN $ in \eqref{loss} can be readily computed to machine precision using reverse mode differentiation \cite{ad}.

The loss function $ L(u;\theta) $ is typically minimized using a stochastic gradient descent algorithm, such as Adam, with a batch of interior and boundary points generated to feed the loss function. The goal of this process is to find a set of neural network parameters $\theta$ that minimize the loss function as much as possible.

It is worth noting that the abstract PDE problem in \eqref{eqn:pde1}-\eqref{eqn:pde2} can easily be extended to time-dependent cases by considering one component of $\bfx$ as a temporal variable. In this case, one or more initial conditions should be included in the PDE system and additional initial condition constraints should be added to the loss function \eqref{loss}.

\subsection{Boundary integral equations}
In this subsection we present a comprehensive overview of the method of formulating BIEs for classical PDE problems. We consider an open, bounded set $\Om\subset\R^d$, with a Lipschitz continuous and piecewise smooth boundary, denoted as $\pa\Om:=\Ga$. The dimension of $\Om$ is denoted by $\dim(\Om)$. The closure of $\Om$ is denoted as $\ol{\Om}$, and its complementary set, which is unbounded, is denoted as $\Om' = \ol{\Om}^c$. Consider the following interior problem ($ P_i $) \eqref{lap_in_eqn}-\eqref{lap_in_bnd},
\begin{empheq}[left=\empheqlbrace]{align} 
\Delta u(\bfx) &= 0 \qquad\qquad \mbox{ in }\Om\label{lap_in_eqn},\\
u(\bfx)	&=	u_0(\bfx)	\qquad \mbox{ on }\Ga\label{lap_in_bnd},
\end{empheq}
and exterior problem ($ P_e $) \eqref{lap_ex_eqn}-\eqref{lap_ex_bnd},
\begin{empheq}[left=\empheqlbrace]{align} 
\Delta u(\bfx) &= 0 \qquad\qquad \mbox{ in }\Om'\label{lap_ex_eqn},\\
u(\bfx)	&=	u_0(\bfx)	\qquad \mbox{ on }\Ga\label{lap_ex_bnd}.
\end{empheq}
It is well known that the fundamental solution $u^\ast$ in this case is given by \cite{katsikadelis2002boundary}:
\begin{equation}
u^\ast(\bfx,\bfy)=
\begin{cases}
	-\frac{1}{2\pi}\ln|\bfx-\bfy| &\quad\mbox{ if } \dim(\Om)=2,\\
	\frac{1}{4\pi|\bfx-\bfy|} &\quad\mbox{ if } \dim(\Om)=3.\\
\end{cases}\label{green_lap}
\end{equation}
By using Green's identity and the fundamental solution of the Laplacian operator, we obtain the following boundary integral representation of the interior problem ($ P_i $):
\begin{equation}
	\int_{\Ga}u^\ast(\bfx,\bfy)\frac{\pa u(\bfx)}{\pa \bfn}-u(\bfx)\frac{\pa u^\ast(\bfx,\bfy)}{\pa \bfn}ds_x=
	\begin{cases}
		u(\bfy)	&\quad\mbox{ if }\bfy\in\Om,\\
		\frac{1}{2}u(\bfy)	&\quad\mbox{ if }\bfy\in\Ga.
	\end{cases}\label{lap_in_formula}
\end{equation}
\begin{remark}
When $ \dim(\Om)=2 $, and the boundary is not smooth at a point  $\bfy\in\Ga $, the factor of $\frac{1}{2}$ in the right-hand side of equation \eqref{lap_in_formula} must be modified to $ \frac{\theta(\bfy)}{2\pi} $, where $ \theta(\bfy) $ is the angle between the two tangent lines at $ \bfy $. In the case where $ \dim(\Om) = 3 $, a similar adjustment must be made, with $\theta(\bfy)$ representing the volume angle at $\bfy$.
\end{remark}
We can get a similar formula for the exterior problem ($ P_e $) under some suitable conditions. 

If we union the interior and exterior problems ($ P_i $) and ($ P_e $), we obtain the global problem ($ P_g $) \eqref{lap_global_eqn}-\eqref{lap_global_bnd}:
\begin{empheq}[left=\empheqlbrace]{align} 
\Delta u(\bfx) &= 0 \qquad\qquad \mbox{ in }\R^d\setminus\Ga\label{lap_global_eqn},\\
u(\bfx)	&=	u_0(\bfx)	\qquad \mbox{ on }\Ga\label{lap_global_bnd}.
\end{empheq}
This is the Laplacian problem in $ \R^d $. Much as before, its solution can be represented by a boundary integral equation on $ \Ga $. The result has been summarized in Theorem \ref{thm_laplace_sol}.
\begin{theorem}\label{thm_laplace_sol}
Let $ u $ be the solution of \eqref{lap_global_eqn}-\eqref{lap_global_bnd}, and $ \Ga $ is smooth. Suppose $ \Delta u $ is continuous on $ \ol{\Om} $ and $ \ol{\Om'} $, and
\[
|u(\bfx)|=O\left(\frac{1}{|\bfx|}\right),\quad |\nabla u(\bfx)|=O\left(\frac{1}{|\bfx|^2}\right),\ \mbox{ as }|\bfx|\to \infty,
\]
then we have the following expression for $ u $:
\begin{equation}
        \int_{\Ga}u^\ast(\bfx,\bfy)\left[\frac{\pa u(\bfx)}{\pa \bfn}\right]-[u(\bfx)]\frac{\pa u^\ast(\bfx,\bfy)}{\pa \bfn}ds_x=
    \begin{cases}
        u(\bfy)	&\quad\mbox{ if }\bfy\in\Om\cup\Om',\\
        \{u(\bfy)\}	&\quad\mbox{ if }\bfy\in\Ga.
    \end{cases}\label{lap_bie}
\end{equation}
\end{theorem}
\begin{remark}
If the boundary is not smooth at $ \bfy\in\Ga $, then the definition of average $ \{\cdot\} $ should be modified to a weighted average with respect to the weighting factor $ \theta(\bfy) $.
\end{remark}

The unknowns in equation \eqref{lap_bie} are $ [\frac{\pa u(\bfx)}{\pa \bfn}] $ and $ [u(\bfx)] $ at points on the boundary $ \Ga $, which are independent of any interior information in $ \mathbb{R}^d\setminus\Ga $. By enforcing equation \eqref{lap_bie} on $\Ga$, we can solve for these unknowns, and then use equation \eqref{lap_bie} to generate the solution of problem ($ P_g $) for any point in $ \mathbb{R}^d $.

\subsection{Neural Operators}\label{sec_nomad}
In this subsection, we first introduce the framework for operator learning and then we present a concise overview of the NOMAD architecture employed in this work \cite{seidman2022nomad}. 

\textbf{Operator Learning:}
Prior to delving into operator learning, it is pivotal that we establish certain notations in order to present a formal definition of the supervised operator learning problem. We denote $C(\mathcal{X}, \R^d)$ as the assembly of continuous functions mapping a set $\mathcal{X}$ to $\R^d$. For instances when $\mathcal{X}\subset\R^n$, we define the Hilbert space as follows,
\[
L^2(\mathcal{X};\R^d)=\left\{f:\mathcal{X}\mapsto\R^d\left|\|f\|^2_{L^2}:=\int_{\mathcal{X}}\|f(x)\|^2_{\R^d}dx<\infty\right.\right\}.
\]

Let's consider a training data-set consisting of $N$ function pairs $(u^i,s^i)$, where each $u^i$ belongs to $C(\mathcal{X};\R^{d_u})$ with $\mathcal{X}\subset\R^{d_x}$ being a compact set, and each $s^i$ resides in $C(\mathcal{Y};\R^{d_s})$ with $\mathcal{Y} \subset\R^{d_y}$ also being a compact set. We presume the existence of a veritable operator $\mathcal{G}:C(\mathcal{X};\R^{d_u})\mapsto C(\mathcal{Y};\R^{d_s})$ such that $\mathcal{G}(u^i)=s^i$, and the $u^i$ are sampled independently and identically from a probability measure on $C(\mathcal{X};\R^{d_u})$.

The primary objective of the supervised operator learning problem is to learn a continuous operator $\mathcal{F}:C(\mathcal{X};\R^{d_u})\mapsto C(\mathcal{Y};\R^{d_s})$ that approximates $\mathcal{G}$. This endeavor involves minimizing the ensuing empirical risk over a class of operators, denoted $\mathcal{F}_\theta$, where $\theta$ is a parameter residing in $\Theta\subset\R^{d_\theta}$,
\[
\mathcal{L}(\theta):=\frac{1}{N}\sum_{i=1}^N\|F_\theta(u^i)-s^i\|^2_{L^2(\mathcal{Y};\R^{d_u})}.
\]

\textbf{Nonlinear Manifold Decoders:}
In this part, we present a concise overview of NOMAD. For more detail about it, please see \cite{seidman2022nomad}. Consider a probability measure $\mu$ on $L^2(\mathcal{X})$ and a mapping $\mathcal{G}:L^2(\mathcal{X})\mapsto L^2(\mathcal{Y})$, with $\mathcal{X}\subset\R^n$. We assume that there exists an $n$-dimensional manifold $\mathcal{M}\subset L^2(\mathcal{Y})$ and an open subset $\mathcal{O} \subset \mathcal{M}$ such that
\[
\mathbb{E}_{u\sim \mu}\left[\inf_{v\in\mathcal{O}}\|\mathcal{G}(u)-v\|^2_{L^2} \right] \leq \epsilon
\]
We refer to this as the operator learning manifold hypothesis. Based on this hypothesis, we introduce a nonlinear decoder $\tilde{D}$ that is parameterized by a deep neural network $f:\R^n\times\mathcal{Y}\mapsto \R$ which jointly takes as arguments $(\beta,y)$, such that
\[
\tilde{D}(\beta,y)=f(\beta,y).
\]
This nonlinear decoder is used to represent target functions. In \cite{seidman2022nomad}, the authors demonstrate the effectiveness of NOMAD through various examples.

\section{Boundary-informed neural operators}\label{sec:BINOMAD}
In this section, we propose the use of NOMAD for solving parametrized PDEs through the re-formulation of PDEs as boundary integral equations. By inferring the unknown boundary data using NOMAD, solutions to these parameterized PDEs can be predicted using only boundary training tasks. As we will demonstrate in the following section, this approach yields satisfactory accuracy and, in some cases, even faster training times compared to traditional numerical solvers.

\subsection{Re-formulation of parameterized PDEs problem}
In this subsection, we use the Laplacian problem as a demonstration to show how to obtain the desired boundary integral equation for the machine learning task in the following subsection. Let us consider a Laplacian problem ($P_1$) by setting a parametrized geometry $ \Ga_t $ in ($P_g$) instead of $\Ga$, where $t$ is the geometric parameter. Due to the Dirichlet boundary condition $ u(\bfx)=u_0(\bfx) $ on $\Ga_t$, the solution $u$ of ($P_1$) is guaranteed to be continuous on $\Ga_t$. Additionally, due to the smoothness of $\Ga_t$, by theorem \ref{thm_laplace_sol}, we arrive at the following boundary integral equation for the representation of $u(\bfy;t)$
\begin{equation}
u(\bfy;t)=\int_{\Ga_t}\left[\frac{\pa u(\bfx;t)}{\pa \bfn}\right]u^\ast(\bfx,\bfy)ds_{x;t},\qquad\mbox{ for }\bfy\in\R^2.\label{exp1_bie}
\end{equation}
It is important to note that the only unknown in \eqref{exp1_bie} is $[\frac{\pa u(\bfx;t)}{\pa \bfn}]$, which is located on the $\Ga_t$.

Since it can be difficult to generate uniformly distributed random points on a general curve or surface, we can only guarantee uniformity on the parameter domain of the curve or surface. In the example ($P_1$), we generate the uniformly random points on the domain of $ t $ for Monte Carlo integration, but a Jacobian must be included in the integrand in this case. Fortunately, the Jacobian is also a function of spatial coordinates $\bfx$ and geometric parameter $t$, so we can set up the unknown $v(\bfx;t)$ as
\[
v(\bfx;t)=\left[\frac{\pa u(\bfx;t)}{\pa \bfn}\right]\left|\frac{ds_{x;t}}{d\tau}\right|,
\]
where $\tau$ is the variable for the boundary representation. Then, the \eqref{exp1_bie} can be written as
\begin{equation}
u(\bfy;t)=\int_{\Ga_t}v(\bfx;t)u^\ast(\bfx(\tau),\bfy)d\tau,\qquad\forall\bfy\in\R^2.\label{exp1_bie1}
\end{equation}
The \eqref{exp1_bie1} is the desired equation that we will use to set up the machine learning task in the following subsection.

\subsection{Data preparation}\label{subsec:data}
From equation \eqref{exp1_bie1}, it is evident that sample points on $\Ga_t$ are required for numerical integration at various values of $t$. The integral equation \eqref{exp1_bie1} already incorporates the PDE \eqref{lap_global_eqn}, thus it is necessary to ensure that the proposed neural network satisfies equation \eqref{lap_global_bnd} in order to fully encode the information contained in equations \eqref{lap_global_eqn} and \eqref{lap_global_bnd}. To accomplish this, sample points, represented by $\bfy_t$ in equation \eqref{lap_global_eqn}, must be obtained on $\Ga_t$.

Based on this analysis, our data preparation will be divided into two distinct phases. Firstly, we will uniformly sample values of $t$ within its domain. For each sampled value of $t$, sample points on $\Ga_t$ will be generated for use in Monte Carlo integration. The coordinates of these points are computed using the equation of $ \Ga_t $ based on the samples of $t$. For the sake of clarity, we refer to this data set as $B$, with a shape of $(N_t, M, b)$, where $N_t$ is the size for the sample $t$, $M$ is the number of points for Monte Carlo integration, and $b$ is the dimension of these points. Additionally, a separate set of observation points on the boundary, represented by $\bfy_t\in\Ga_t$, will be generated randomly to form the loss function. We refer to this data set as $T$, with a shape of $(N_y, a)$, where $N_y$ is the sample size for $\bfy_t$ and $a$ is the dimension of the sample $\bfy_t$.

\subsection{Neural network design}
In Section \ref{sec_nomad}, we briefly introduced NOMAD and in this subsection, we will delve deeper into its neural network architecture for solving ($P_g$). The goal of the proposed NOMAD is to infer the solution $v(\bfx;t)$ in equation \eqref{exp1_bie1} by means of the boundary condition in equation \eqref{lap_global_bnd}. To accomplish this, we designed our neural network to take data-sets $B$ and $T$ as input and output the corresponding $v(\bfx;t)$. To begin, we pre-defined a hyper-parameter, $p$, as the number of features to be extracted from the encoder and decoder. Next, data-set $T$ is passed through the encoder, resulting in an output of shape $(N_y, p)$. Similarly, data-set $B$ is passed through the decoder, resulting in an output of shape $(N_t, M, p)$. The selection of encoder and decoder networks is dependent on the problem at hand. In our experiment, we employed a fully connected network as the encoder and a Fourier feature network as the decoder.

\subsection{Loss function}
The objective of the loss function in this task is to ensure that the boundary condition specified in equation \eqref{lap_global_bnd} is met through the solution representation outlined in equation \eqref{exp1_bie1}. In other words, we aim to ensure that the output of the neural network, $v(\bfx;t)$, satisfies the following equation for all $\bfy_t\in\Ga_t$:
\begin{equation}
u_0(\bfy_t)=\int_{\Ga_t}v(\bfx;t)u^\ast(\bfx,\bfy_t)d\bfx,\quad\forall\bfy_t\in\Ga_t.\label{loss1}
\end{equation}
To this end, we set the loss function as
\begin{equation}
L=\frac{1}{N_yN_t}\sum_{i=1}^{N_y}\sum_{j=1}^{N_t}\left|\frac{1}{M}\sum_{k=1}^Mv(\bfx_k;t_j)u^\ast(\bfx_k,\bfy_{i;t})-u_0(\bfy_{i;t})\right|^2.\label{loss2}
\end{equation}

Note that the function $u^\ast(\bfx,\bfy)$ in equation \eqref{loss1} is singular at the point $\bfx=\bfy$. To address this, a threshold value, denoted as $\beta$, is established. Any evaluation results of $u^\ast(\bfx,\bfy)$ that are greater than $\beta$ or are not a number (NaN) will be truncated to $\beta$. This helps to ensure that the function remains well-defined and computationally manageable.

\section{Numerical Experiments}\label{sec:numerical}
In this section, we will validate the effectiveness of the proposed algorithm through a series of comprehensive numerical experiments. We will detail specific hyper-parameter setups in each subsequent subsection, but will utilize the default settings outlined in Table \ref{default_setup} for certain hyper-parameters across all experiments.
\begin{table}
 \caption{Default Experiment Set up}\label{default_setup}
  \centering
  \begin{tabular}{ll}
    \toprule
    Name     & Value \\
    \midrule
    $\beta$               & $ 100,000 $ \\
    $ p $       & $ 100 $ \\
    Training steps      &   $ 200,000 $    \\
    Activation function & GeLU    \\
    Method to initialize the neural network & Xavier     \\
    Optimizer   &   Adam      \\
    Learning rate  & $ 10^{-3} $      \\
    Learning rate decay period     &   $ 20,000 $     \\
    Learning rate decay rate  & $ 0.95 $      \\
    \bottomrule
  \end{tabular}
  \label{tab:table}
\end{table}

To evaluate the model's accuracy, we employ the relative $ l^2 $ error, calculated over a set of points $ \{\bfx_i\}_{i=1}^N $, defined as:
\[
err = \frac{\sum_{i=1}^{N}|u_{pred}(\bfx_i)-u_{true}(\bfx_i)|^2}{\sum_{i=1}^{N}|u_{true}(\bfx_i)|^2},
\]
for a given geometric parameter $t$.
\subsection{2D Laplace equation}\label{subsec:ex_lap}
We start our numerical experiments exhibition with the 2D Laplace equation as shown in ($P_1$), which is commonly used by a showcase in many computational mathematics papers to illustrate the concepts.

Utilizing the Green's function of the 2D Laplace operator as shown in equation \eqref{green_lap}, we proceed to establish the BIE to be solved, as described in equation \eqref{exp1_bie1}. Specifically, we set the initial condition as $u_0(\bfx)=e^x\sin(y)$ in ($P_1$), where $\bfx=(x,y)$. The boundary $\Ga_t$ is parametrized by a geometric parameter $t\in[1,2]$, with a polar coordinate representation given by the following equation:
\begin{equation}
r(\alpha;t)=1+0.2(\sin(3\alpha)+t\sin(4\alpha)+\sin(6\alpha)+\cos(2\alpha)+\cos(5\alpha)),\quad\alpha\in[0,2\pi).\label{lap_ex_geo}
\end{equation}
To generate random points on $\Ga_t$, we will generate uniformly random points in $[0,2\pi)$ and then map them to polar coordinates through equation \eqref{lap_ex_geo}. The dimension $d=2$ is obvious.

In this experiment, we employed a fully-connected neural network with $ 3 $ hidden layers, each with a size of $100$, as the encoder. The decoder is designed as a Fourier feature network with standard Gaussian initialization \cite{tancik2020fourier}. The number of points for Monte Carlo integration $M$, as mentioned in subsection \ref{subsec:data}, is $3,000$. For each batch of data in training, we randomly generated $10$ geometric parameters and $100$ observation points, i.e., $ N_t=10 $ and $ N_y=100 $.

\begin{figure}
\centering
\subfigure[Results for $t=1.15$. Relative $l^2$ error: $ 2.85\% $.]{
\begin{minipage}[b]{1\textwidth}
  \includegraphics[width=1\textwidth]{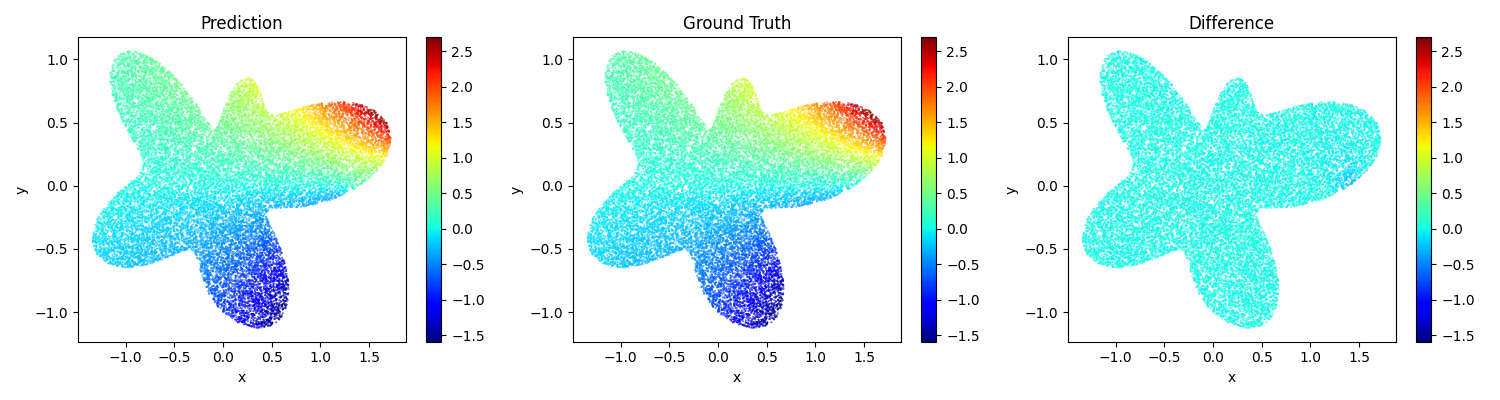} \\
  \end{minipage}
}

\subfigure[Results for $t=1.35$. Relative $l^2$ error: $ 2.87\% $.]{
\begin{minipage}[b]{1\textwidth}
  \includegraphics[width=1\textwidth]{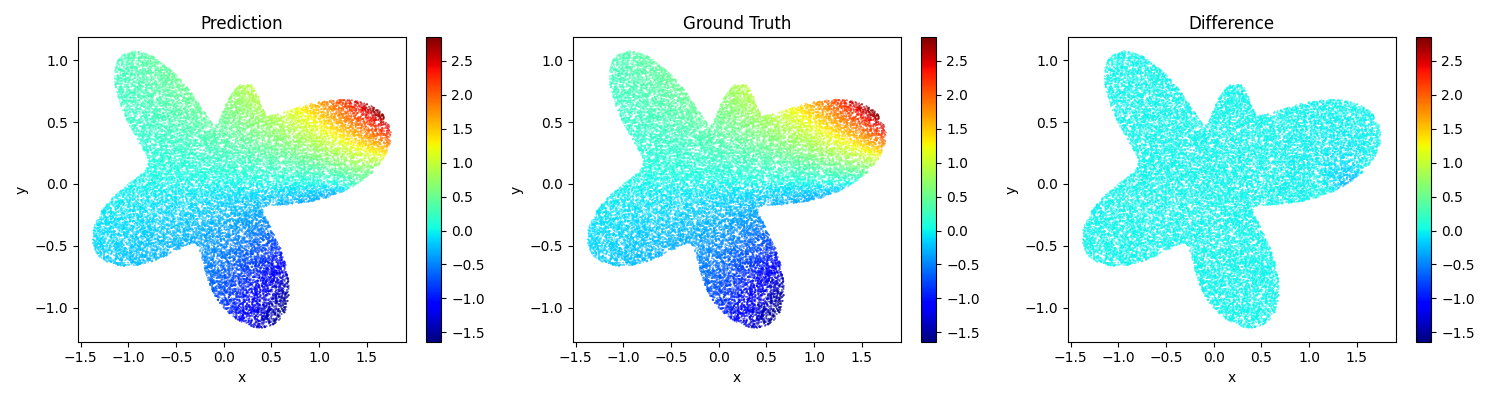} \\
  \end{minipage}
}

\subfigure[Results for $t=1.45$. Relative $l^2$ error: $ 3.00\% $.]{
\begin{minipage}[b]{1\textwidth}
  \includegraphics[width=1\textwidth]{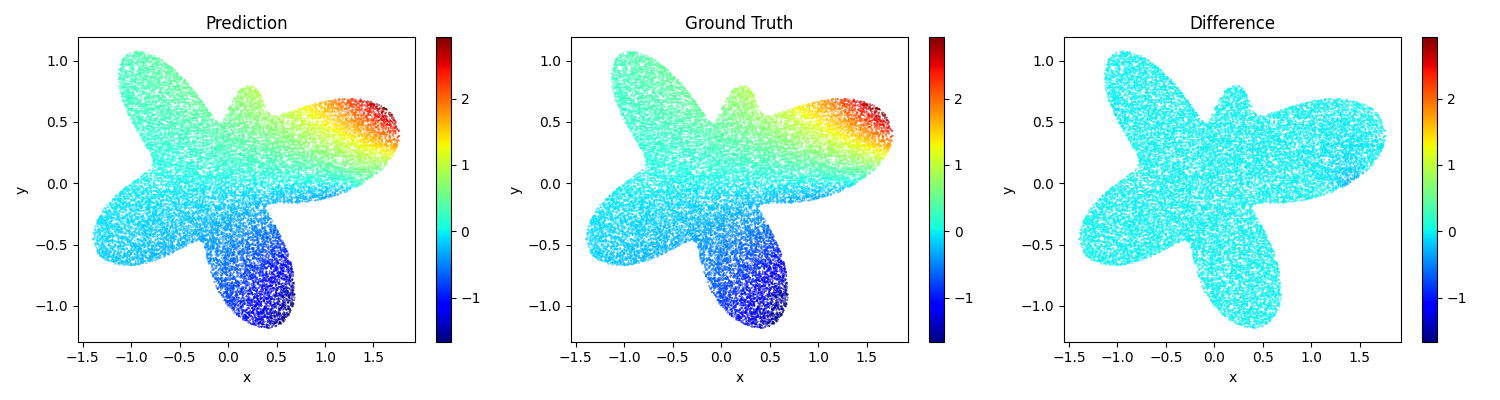} \\
  \end{minipage}
}

\caption{
{\em Learning the boundary operator of 2D parametric Laplace equation:} Comparisons between the model predictions and the ground truth at different time $t$. (a) The results at $t=1.15$ with a relative error of $ 2.85\% $.  (b) The results at $t=1.15$ with a relative error of $ 2.87\% $. (c)  The results  at $t=1.45$ with a relative error of $ 3.00\% $. 
Detailed hyper-parameter sweep studies can be found in Appendix \ref{appx1}.}
\label{fig:exp0_lap}
\end{figure}

The results of this experiment are presented in Fig. \ref{fig:exp0_lap}. To provide a clear illustration of the boundary, the solution is only displayed within the boundary $\Ga_t$ instead of whole $ \R^2 $. We note that in the last experiment, an exterior solution is necessary, and we will show an exterior solution therein. In this experiment, we predicted the solution at $t= 1.15 $, $ t=1.35 $, and $ t=1.45 $. These values are chosen randomly and do not hold any specific significance. The relative errors for each sample of $ t $ are: $ 2.85\% $, $ 2.87\%$, and $ 3.00\% $, respectively.

Hyper-parameter sweep studies were conducted and presented in Appendix \ref{appx1}. The studies reveal that as the network architecture and $p$ increase, there is a trend towards a decrease in relative error. However, this trend is not always evident, and there are instances where larger architecture and larger $p$ result in larger relative errors. One reason for this is over-fitting, a common phenomenon in machine learning. Another reason is that the Monte Carlo integration for integrals on boundaries with singularities limits the accuracy of the network's output, thereby contaminating the solution. Utilizing special quadrature rules can address this issue, but it incurs a high computational cost due to the need to generate a unique rule for each geometry $\Ga_t$. While the Monte Carlo integration used in this study is subject to accuracy limitations, it is easy to implement for complex geometries and allows for multi-GPU training for large problems.

\subsection{2D Bi-harmonic equation}\label{subsec:ex_bih}
In this section, we examine a 2D bi-harmonic equation as outlined in equations \eqref{bih_in_eqn}-\eqref{bih_neu_bnd}.
\begin{empheq}[left=\empheqlbrace]{align} 
\Delta^2 u(\bfx) &= 0 \qquad\qquad \mbox{ in }\R^2\setminus\Ga_t\label{bih_in_eqn},\\
u(\bfx)	&=	u_0(\bfx)	\qquad \mbox{ on }\Ga_t\label{bih_dir_bnd},\\
\frac{\pa u(\bfx)}{\pa\bfn}	&=	\frac{\pa u_0(\bfx)}{\pa\bfn}	\qquad \mbox{ on }\Ga_t\label{bih_neu_bnd}.
\end{empheq}
The bi-harmonic operator $\Delta^2$ in \eqref{bih_in_eqn} is a fourth-order differential operator. In a traditional PINN framework, computing the gradient four times per dimension would result in a large computational graph. By utilizing the BIE technique, however, the order of derivatives is reduced to first order as shown in \eqref{exp2_bie}. Additionally, the proposed method only requires training on the boundary points, resulting in an extremely efficient machine learning solver.

The BIE for the bi-harmonic problem \eqref{bih_in_eqn}-\eqref{bih_neu_bnd} is as follows \cite{katsikadelis2002boundary}:
\begin{equation}
u(\bfy) =-\int_{\Ga_t}\frac{\pa v(\bfx;t)}{\pa\bfn}u^\ast(\bfx,\bfy)+v(\bfx;t)\frac{\pa u^\ast(\bfx,\bfy)}{\pa\bfn_x}ds_{x;t}\quad\forall \bfy\in\R^2,\label{exp2_bie}
\end{equation}
where the Green's function is defined as:
\[
u^\ast(\bfx,\bfy)=\frac{1}{8\pi}|\bfx-\bfy|^2\ln|\bfx-\bfy|,
\]
and $v(\bfx;t)$ represents the unknown function on the boundary that needs to be determined.

\begin{figure}
\centering
\subfigure[Results for $t=1.15$. Relative $l^2$ error: $ 1.08\% $.]{
\begin{minipage}[b]{1\textwidth}
  \includegraphics[width=1\textwidth]{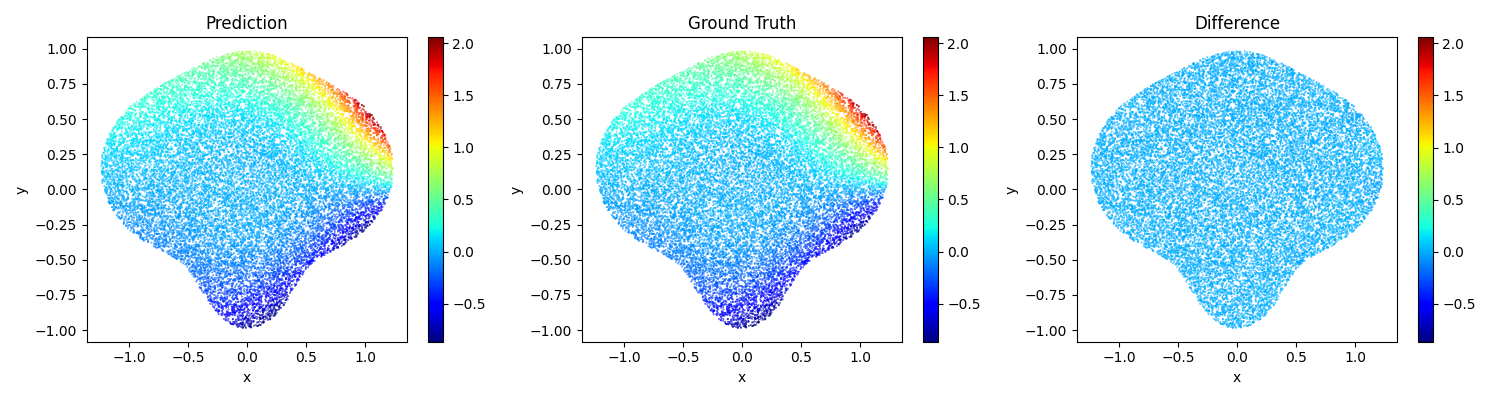} \\
  \end{minipage}
}
\subfigure[Results for $t=1.35$. Relative $l^2$ error: $ 0.90\% $.]{
\begin{minipage}[b]{1\textwidth}
  \includegraphics[width=1\textwidth]{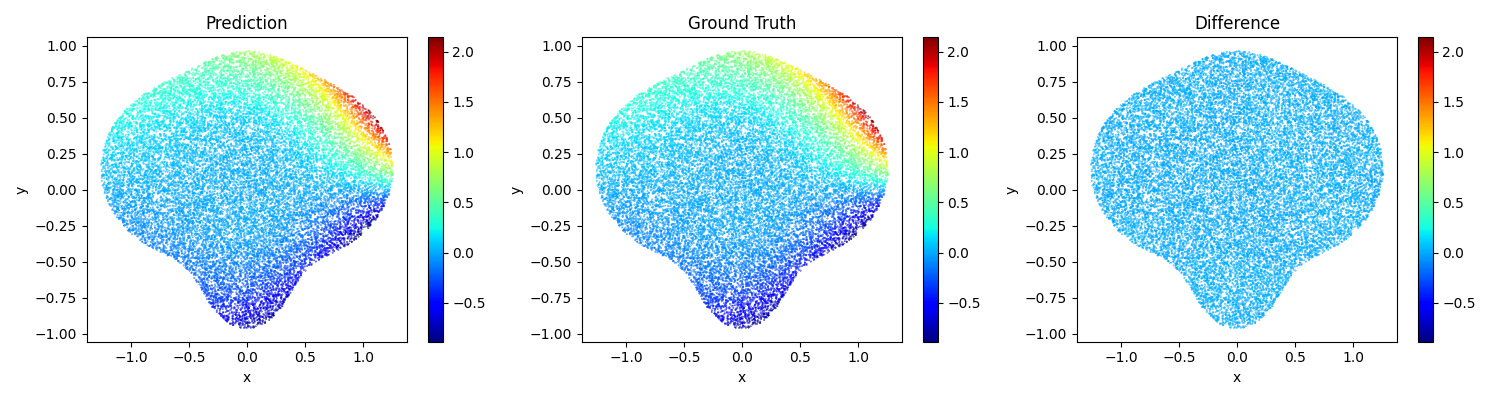} \\
  \end{minipage}
}
\subfigure[Results for $t=1.45$. Relative $l^2$ error: $ 0.77\% $.]{
\begin{minipage}[b]{1\textwidth}
  \includegraphics[width=1\textwidth]{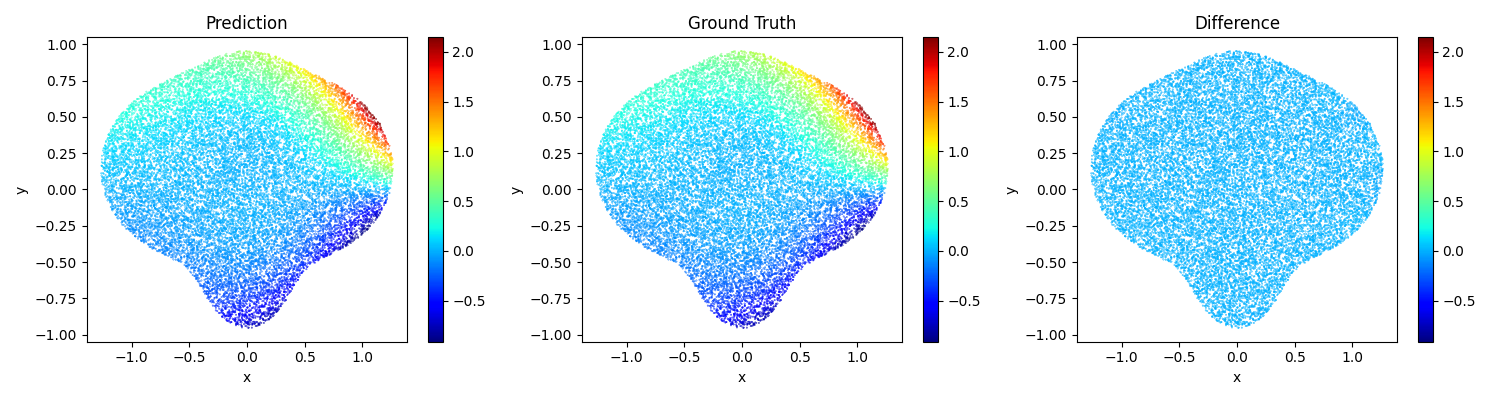} \\
  \end{minipage}
}
\caption{Results of the 2D Parametrized Bi-Harmonic Equation. Depictions of the parametrized geometry as given in \eqref{bih_ex_geo} for varying values of $t$: (a) $t=1.15$ with a relative error of $ 1.08\% $, (b) $t=1.35$ with a relative error of $ 0.90\% $, and (c) $t=1.45$ with a relative error of $ 0.77\% $.}
\label{fig:exp1_bih}
\end{figure}

In this example, the geometry $\Ga_t$ is defined by the following equation:
\begin{equation}
    r(\alpha;t)=1+0.1(\sin(\alpha)+t\cos(2\alpha)+\sin(3\alpha)+\cos(4\alpha)).\label{bih_ex_geo}
\end{equation}
The same network structures and the number of points for each data set were used as in the previous example. The synthetic solution $u_0(\bfx)=(x^2+y^2)e^x\sin(y)$ was chosen. The results are displayed in Figure \ref{fig:exp1_bih}. As before, the solution was predicted at $t = 1.15$, $t = 1.35$, and $t = 1.45$. The relative errors for each sample of $t$ are $1.08\%$, $0.90\%$, and $0.77\%$, respectively.

\subsection{3D Helmholtz equation}\label{subsec:ex_hel}
In this subsection, we utilize our proposed method to solve the Helmholtz equation in the presence of a parameterized obstacle. The governing equation is given by:
\begin{empheq}[left=\empheqlbrace]{align} 
\Delta u(\bfx) + k^2u(\bfx) &= 0 \qquad\qquad \mbox{ in }\R^3\setminus\overline{\Om}_t\label{har_eqn},\\
u(\bfx)	&=	u_0(\bfx)	\qquad \mbox{ on }\Ga_t\label{har_bnd},
\end{empheq}
where $\Om_t$ is a bounded closed subset in $\R^3$, $\Ga_t=\pa\Om_t$, and $k\in\R$ is the wave number, which cannot be a Laplacian eigenvalue. This problem, given by \eqref{har_eqn}-\eqref{har_bnd}, is an exterior problem that describes the scattering of time-harmonic acoustic or electromagnetic waves by a penetrable in-homogeneous medium of compact support and by a bounded impenetrable obstacle ($\Om_t$ in \eqref{har_eqn}). People are often interested in the wave distribution around the obstacle as well as its limit when the distance goes to infinity, which is known as the far field, as we will define below. Many works have focused on solving \eqref{har_eqn}-\eqref{har_bnd} numerically, with a summary provided in \cite{colton1998inverse}. Since we are concerned with the exterior solution, which is usually unbounded, traditional numerical methods are not suitable for this situation, as well as traditional PINNs. In \cite{aussal2022computing}, the authors thoroughly investigate how to use the boundary element method to solve this unbounded problem.

The Green's function for \eqref{har_eqn}-\eqref{har_bnd} is given by
\[
u^\ast(\bfx,\bfy)=\frac{1}{4\pi}\frac{e^{ik|\bfx-\bfy|}}{|\bfx-\bfy|},
\]
and the corresponding BIE is:
\begin{equation}
u(\bfy) =\int_{\Ga_t}v(\bfx;t)u^\ast(\bfx,\bfy)ds_{x;t}\quad\forall \bfy\in\R^3,\label{exp3_bie}
\end{equation}
where $ v(\bfx;t) $ is the unknown function on $\Ga_t$ to be solved. Although \eqref{har_eqn} is only defined outside $\Om_t$, we extend this solution to the entire $\R^3$. This is because, otherwise, we would need to take the average for the left-hand side in \eqref{exp3_bie} on $\Ga_t$, which would break its unity and make it more difficult to implement. One important note is that the original problem, \eqref{har_eqn}-\eqref{har_bnd}, can be solved in the real domain if there are no complex value boundary conditions. However, the representation \eqref{exp3_bie} requires us to work in the complex value domain. Therefore, we will see that our neural network's output is complex.

The far-field pattern, $u_\infty$, is a crucial aspect of scattering analysis as it allows us to infer the properties of the obstacle, such as its shape and location. It is defined as a function on the unit sphere, $\S^2$, and is given by the following equation:
\[
u_\infty(\overline{\bfx};t)=\frac{1}{4\pi}\int_{\Ga_t}e^{-ik\overline{\bfx}\cdot\bfx}v(\bfx;t)ds_{x;t},\quad\forall\overline{\bfx}\in\S^2. 
\]
This equation describes the wave distribution around the obstacle as well as its limit when the distance goes to infinity, which is commonly referred to as the far-field. The computation of $u_\infty$ is essential in determining the properties of the obstacle and is widely used in scattering analysis.

\begin{figure}[htbp]
\centering
\begin{minipage}[t]{0.48\textwidth}
\centering
\includegraphics[width=1\textwidth]{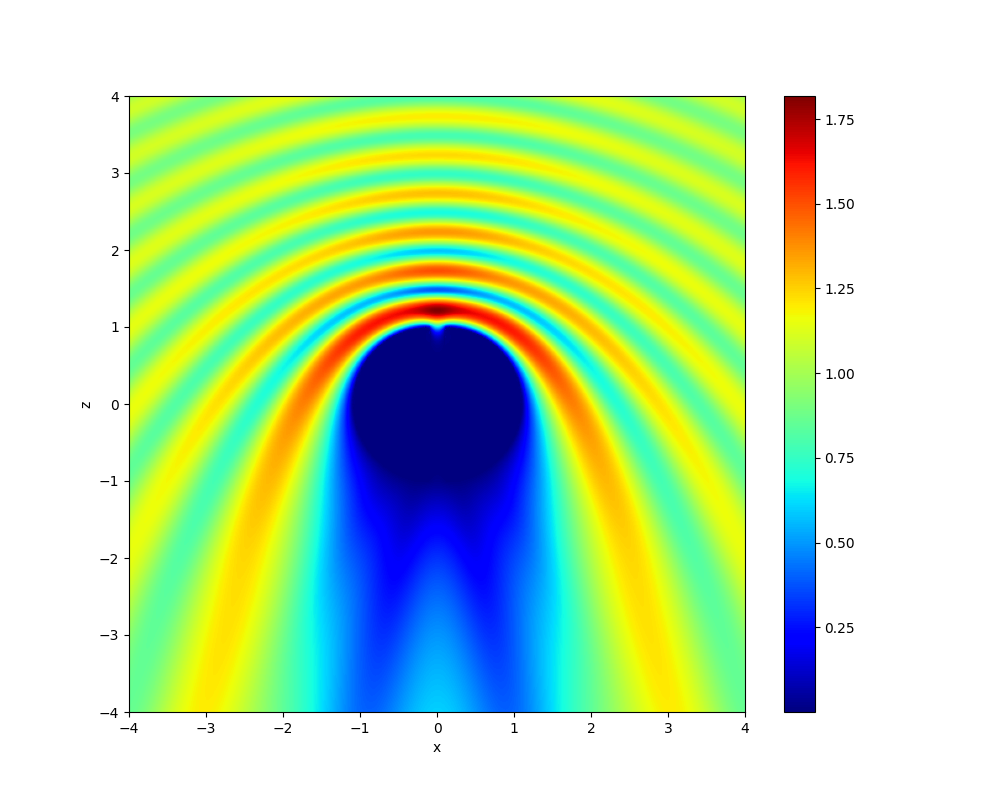}
\caption*{(a) Total field for $t=0.1$.}
\end{minipage}
\begin{minipage}[t]{0.48\textwidth}
\centering
\includegraphics[width=1\textwidth]{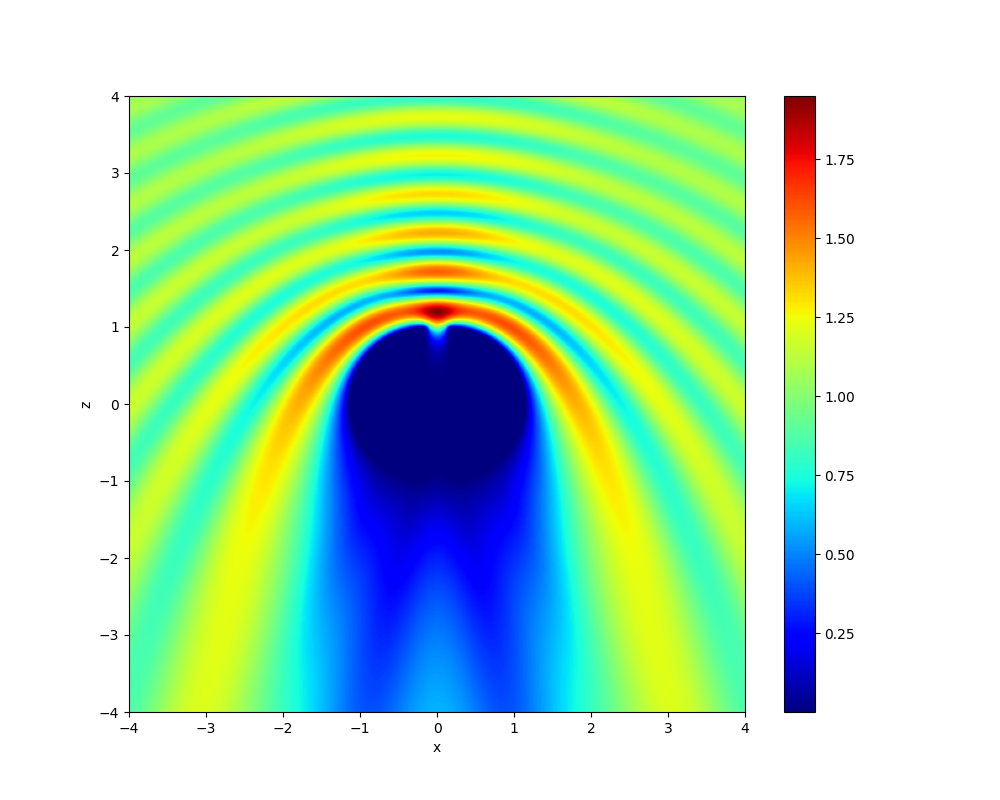}
\caption*{(b) Total field for $t=0.15$.}
\end{minipage}
\begin{minipage}[t]{0.48\textwidth}
\centering
\includegraphics[width=1\textwidth]{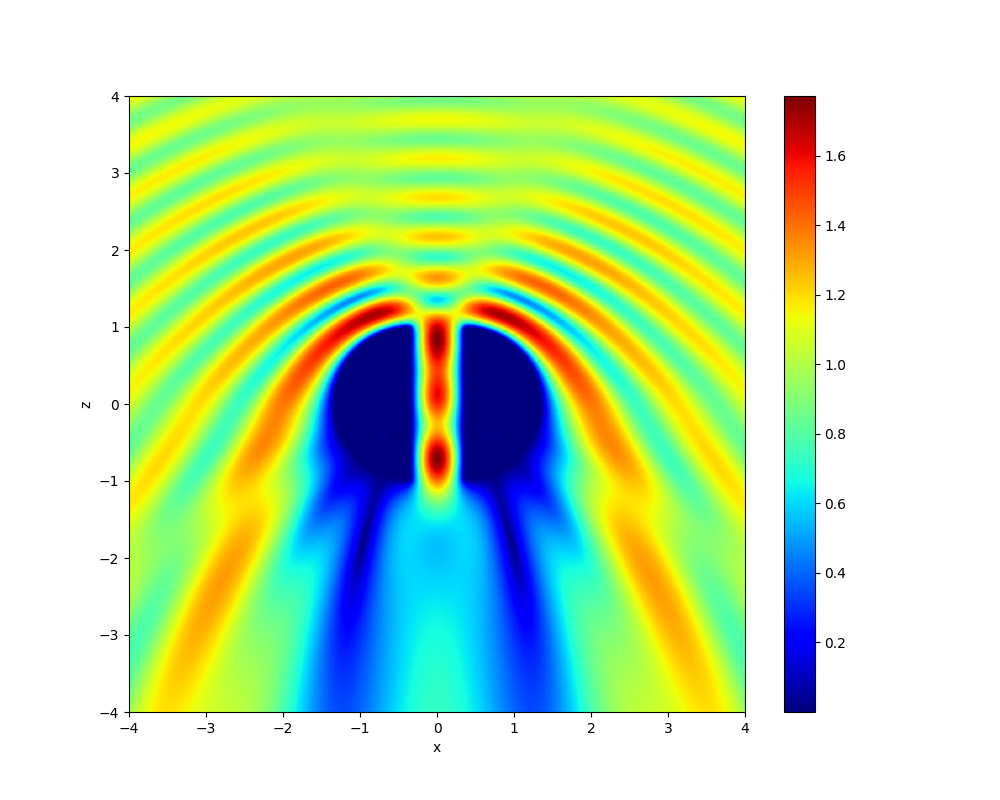}
\caption*{(c) Total field for $t=0.35$.}
\end{minipage}
\begin{minipage}[t]{0.48\textwidth}
\centering
\includegraphics[width=1\textwidth]{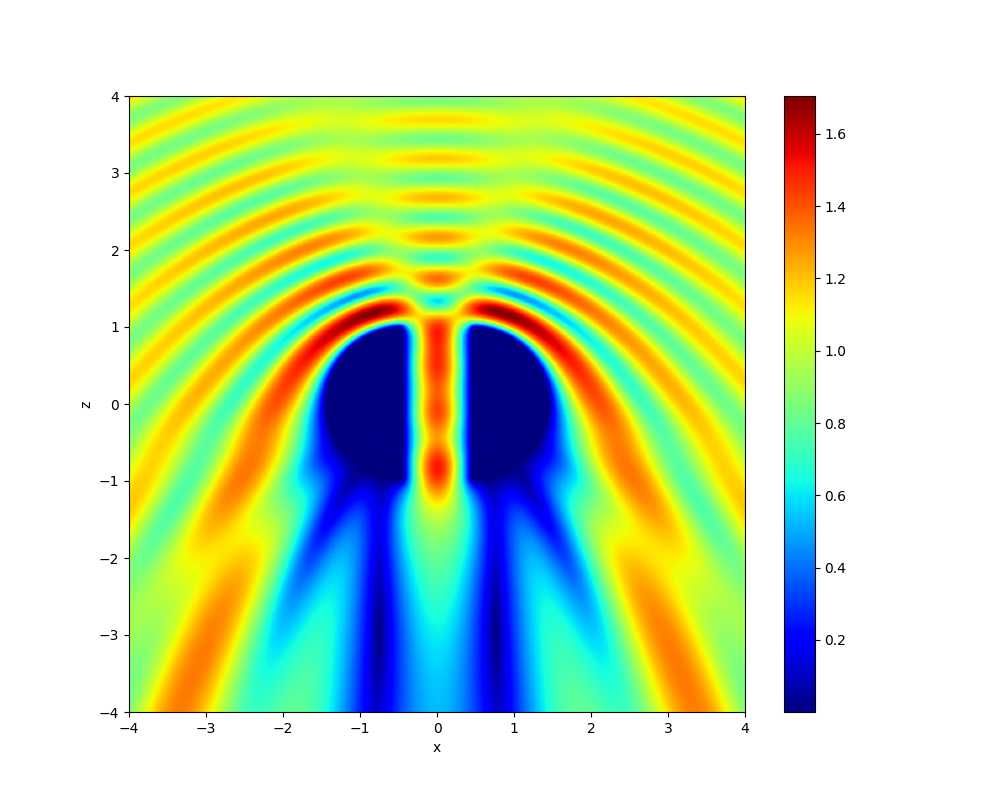}
\caption*{(d) Total field for $t=0.45$.}
\end{minipage}
\caption{Predicted total field for the parameterized obstacle on the slice at $y=0$ with different distances $t =0.1, 0.15, 0.35, 0.45$, respectively.
}
\label{fig:exp3_scattering}
\end{figure}

Our proposed algorithm was able to accurately reproduce the results of the experiment in \cite{aussal2022computing}. We considered the same scenario of a scattering of a plane wave $u^i(r,\theta)=e^{ikr\cos(\theta)}$ by two half-spheres of radius $1$ centered at $(0,0,\pm t)$, for $t\in[0,0.5]$. As in the original experiment, when $t=0$, the problem degenerates to a scattering of the incident wave by a unit sphere. The geometries can be identified from Fig. \ref{fig:exp3_scattering}. Our results were in good agreement with the results reported in \cite{aussal2022computing}, demonstrating the effectiveness of our proposed algorithm in solving this type of scattering problem.

To begin, we applied our proposed algorithm to solve for $v(\bfx;t)$ and generate the far-field pattern $u_\infty(\overline{\bfx};t)$. We compared our results to reference solutions generated by BEMpp, a Python library for numerical boundary elements, to calculate the relative $l^2$ errors. In this example, the dimension is clearly $3$. We set the number of points for Monte Carlo integration to $M=56,000$, the number of observation points to $N_y=1,880$, and the number of parameters to $N_t=2$. The wave number was set to $k=2\pi$.

\begin{figure}[htbp]
\centering
\begin{minipage}[t]{0.48\textwidth}
\centering
\includegraphics[width=1\textwidth]{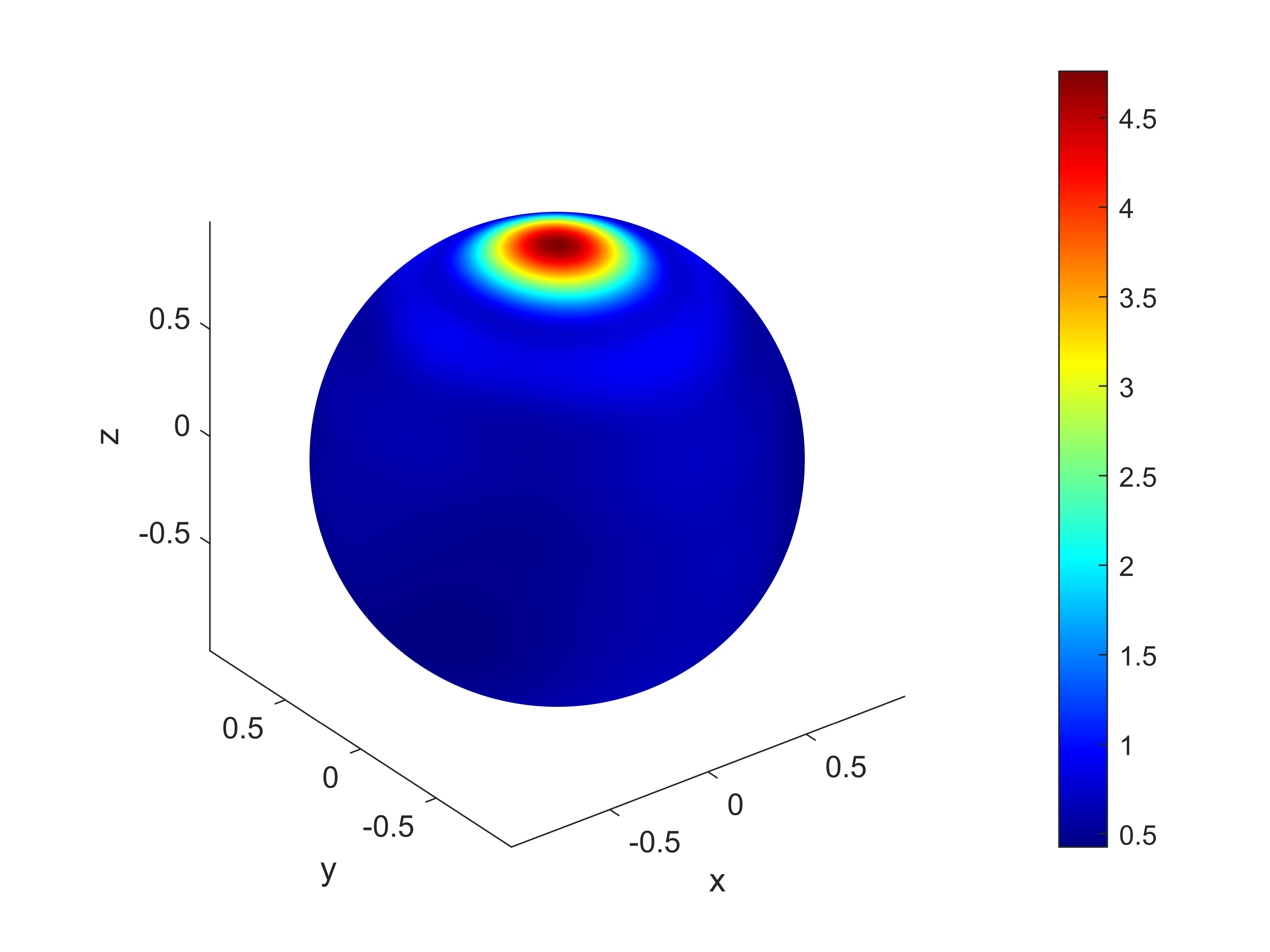}
\caption*{(a) Far-field $u_\infty$ for $t=0.1$.}
\end{minipage}
\begin{minipage}[t]{0.48\textwidth}
\centering
\includegraphics[width=1\textwidth]{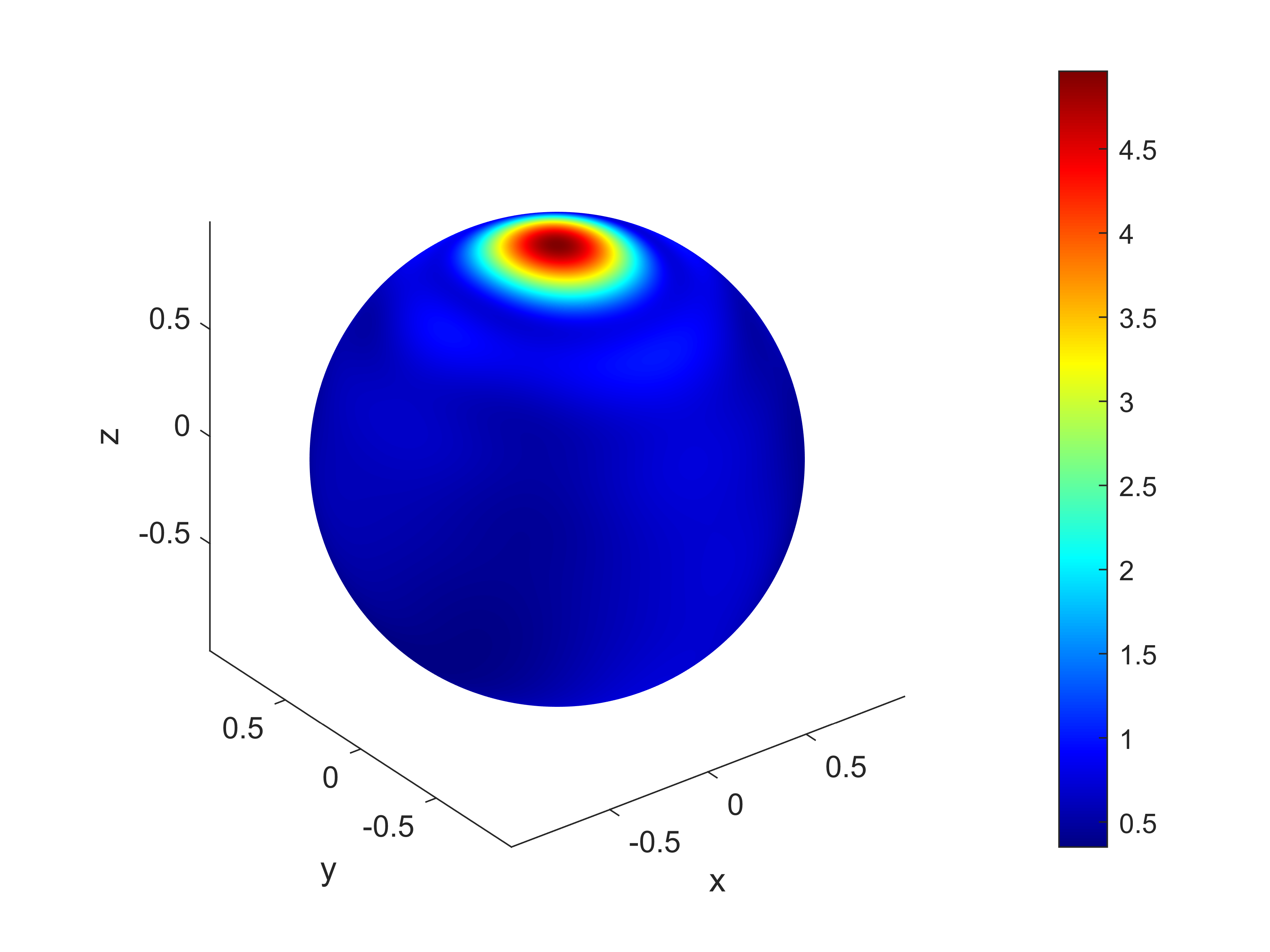}
\caption*{(b) Far-field $u_\infty$ for $t=0.15$.}
\end{minipage}
\begin{minipage}[t]{0.48\textwidth}
\centering
\includegraphics[width=1\textwidth]{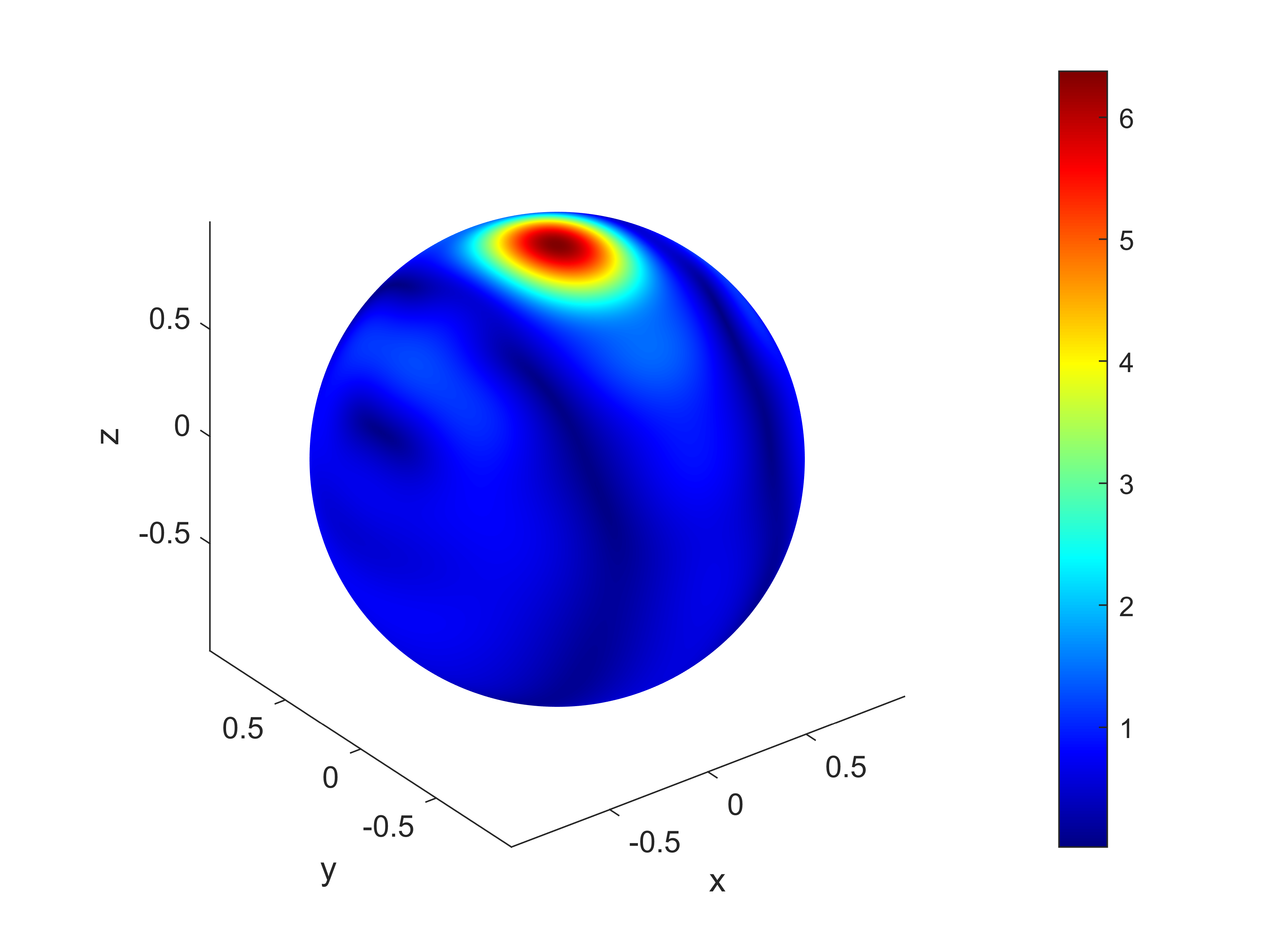}
\caption*{(c) Far-field $u_\infty$ for $t=0.35$.}
\end{minipage}
\begin{minipage}[t]{0.48\textwidth}
\centering
\includegraphics[width=1\textwidth]{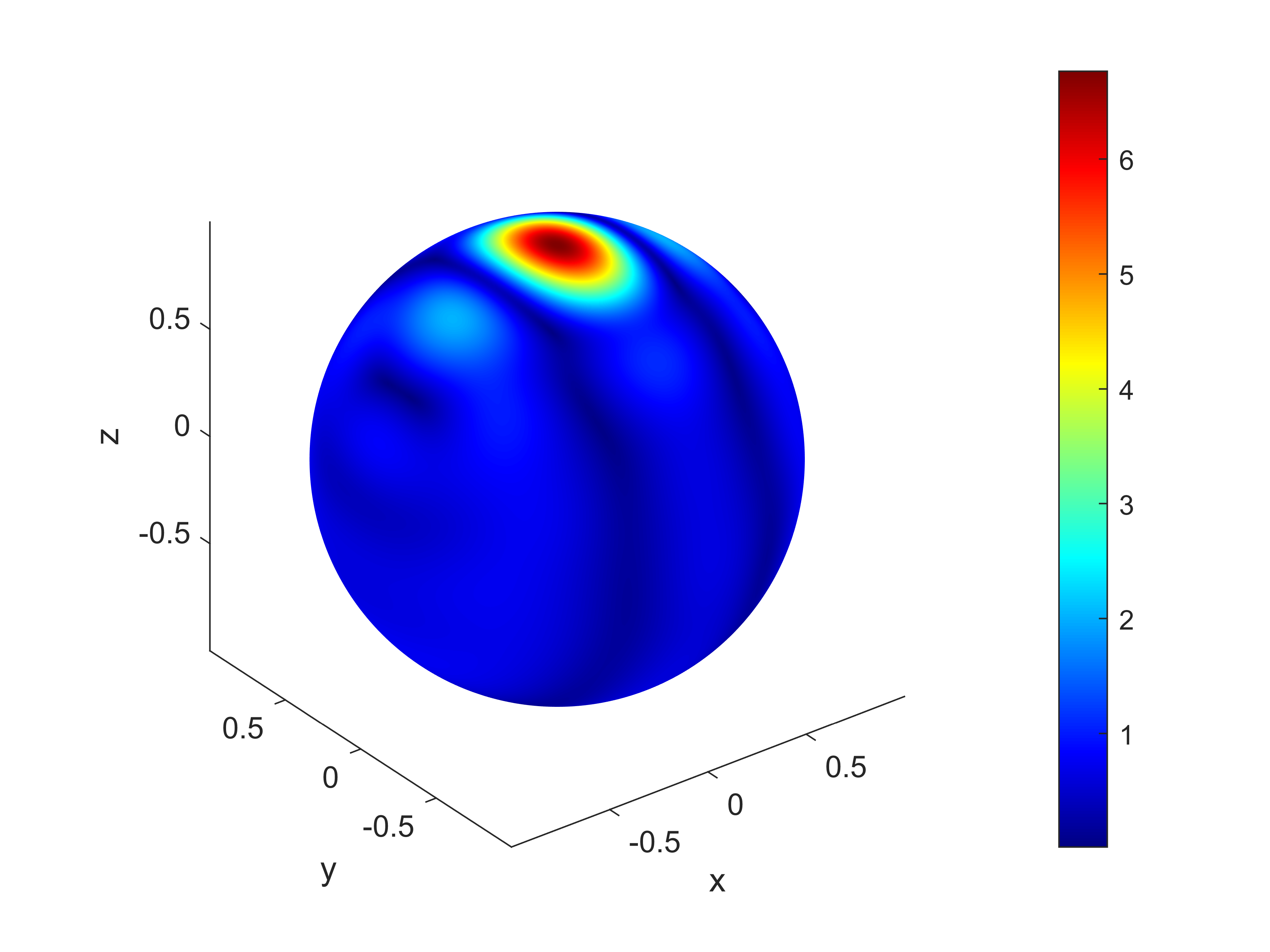}
\caption*{(d) Far-field $u_\infty$ for $t=0.45$.}
\end{minipage}
\caption{ {\em Predicted far-fields for the Parameterized Obstacle:}  Far-field predictions with two hemispheres at different parameterized distances of t. (a) The results  at $t=0.1$, with a relative error of $3.56\%$.  (b) The results  at $t=0.15$, with a relative error of $3.72\%$, (c) The results  at $t=0.35$, with a relative error of $4.18\%$, and (d) $t=0.45$ with a relative error of $3.10\%$.
}
\label{fig:exp3_far}
\end{figure}

We evaluated the accuracy of the far-field solution $u_\infty(\overline{\bfx};t)$ at $t=0.1,\ 0.15,\ 0.35,\ 0.45$. The corresponding relative errors were found to be $3.56\%$, $3.72\%$, $4.18\%$, and $3.10\%$, respectively. The choice of using the far-field as a benchmark for accuracy is motivated by its defined nature on a closed domain $\S^2$. This is in contrast to the scattering field, which is defined on an unbounded set, making it difficult to verify its accuracy. Furthermore, the far-field pattern is known to be more relevant in many real-world applications compared to the scattering field. The far-field patterns generated by the proposed method can be seen in Fig. \ref{fig:exp3_far}.

Subsequently, we employed the solution $v(\bfx;t)$ to generate the total field on the unbounded domain $\R^3\setminus\Om_t$. This was done to evaluate the capability of the proposed method in generating solutions for unbounded domains. Since the total field is distributed all around $\R^3\setminus\overline{\Om}_t$, it is impossible to visualize the entire total field. Instead, we will plot the total field at $y=0$ limited in the box $(x,z)\in[-4,4]^2$. The plots of the cases when $t=0.1,\ 0.15,\ 0.35,\ 0.45$ are shown in Fig. \ref{fig:exp3_scattering}.

\section{Conclusions}\label{sec:conclusion}
In this work, we propose a machine learning-based solver that utilizes the BIEs and PINNs to solve PDEs. By reformulating PDEs into BIEs, the PINNs method can determine the unknown potentials on the boundary by using boundary conditions, and subsequently generate solutions anywhere in the domain. This approach has several advantages over traditional PINNs. Firstly, it reduces the training workload by only utilizing sets of points on the boundary, making it particularly suitable for complex geometries. Additionally, by generating solutions through boundary information, it enables the solution of unbounded domain problems which are not possible with traditional PINNs. Furthermore, BIEs lower the order of derivatives in PDEs, simplifying the computational graph and accelerating the training process. The numerical examples presented in this work demonstrate that this is a promising approach for certain types of PDEs.

However, it must be noted that this approach also has certain limitations. Firstly, it is only applicable to a limited class of problems and is not suitable for nonlinear problems or linear problems with variable coefficients. Secondly, it requires Monte Carlo integration to compute singular integrals, which limits the accuracy of the method. Therefore, further research is required in this area. One direction of research could be to investigate algorithms for nonlinear problems or linear problems with variable coefficients, drawing inspiration from existing boundary element methods for nonlinear problems. Additionally, the accuracy of the integration of singular integrals on a point cloud should be studied, and better algorithms can be expected in this area. Finally, the proposed algorithm can be used in inverse problems such as inverse scattering problems, but further research is needed to develop appropriate ways to represent the underlying geometry.

\section*{Acknowledgments}
This was supported in part by the US Department of Energy under the Advanced Scientific Computing Research program (grant DE-SC0019116) and the US Air Force (grant AFOSR FA9550-20-1-0060).

%Bibliography
\bibliographystyle{unsrt}  
\bibliography{references}  

\section{Appendix}
\subsection{Hyper-paramemters sweep}\label{appx1}
To further evaluate the effectiveness of our method, we conducted hyper-parameter sweep studies for the Laplace example in subsection \ref{subsec:ex_lap} to quantify its predictive accuracy for different neural network architectures, different values of $p$, and with or without Fourier encoding in the decoder. Specifically, we varied the number of layers to $1$, $3$, $5$, and $7$ (referred to as "Layers" in the tables below), varied the layer size to $10$, $50$, $100$, and $150$ (referred to as "Neurons" in the tables below), and varied $p$, the number of features extracted from the encoders and decoders, to $10$, $50$, $100$, and $150$. In the results presented below, we will display the relative $l^2$ error for $t=1.15$.

\subsubsection{Hyper-parameters sweep with Fourier feature decoder}
We present the results for the Fourier feature decoder first, which can be found in Table \ref{apdx_sweep_f10}-\ref{apdx_sweep_f150}.

\begin{table}
 \caption{Hyper-parameters sweep with Fourier feature decoder, $p=10$.}\label{apdx_sweep_f10}
  \centering
  \begin{tabular}{c|cccc}
    \toprule
    \diagbox{Layers}{Neurons}     & $10$  &     $50$    &   $100$   &   $150$  \\
    \midrule
    $1$ &   $49.48\%$   &   $31.88\%$     &     $43.84\%$     &   	$40.31\%$  \\
    $3$ &  $13.42\%$    &	$3.44\%$      &   	$2.56\%$      &   	$3.19\%$   \\
    $5$ &   $9.93\%$    &	$2.80\%$      &   	$4.49\%$      &     $3.30\%$   \\
    $7$ &   $4.68\%$    &	$3.19\%$      &     $2.70\%$      &     $2.11\%$   \\
    \bottomrule
  \end{tabular}
\end{table}

\begin{table}
 \caption{Hyper-parameters sweep with Fourier feature decoder, $p=50$.}\label{apdx_sweep_f50}
  \centering
  \begin{tabular}{c|cccc}
    \toprule
    \diagbox{Layers}{Neurons}     & $10$  &     $50$    &   $100$   &   $150$  \\
    \midrule
    $1$ &   $5.96\%$    &   $3.77\%$      &     $2.16\%$      &   	$2.07\%$   \\
    $3$ &   $3.07\%$    &	$2.22\%$      &   	$1.58\%$      &   	$2.11\%$   \\
    $5$ &   $3.01\%$    &	$1.78\%$      &   	$2.40\%$      &     $2.85\%$   \\
    $7$ &   $4.58\%$    &	$3.43\%$      &     $2.34\%$      &     $2.51\%$   \\
    \bottomrule
  \end{tabular}
\end{table}

\begin{table}
 \caption{Hyper-parameters sweep with Fourier feature decoder, $p=100$.}\label{apdx_sweep_f100}
  \centering
  \begin{tabular}{c|cccc}
    \toprule
    \diagbox{Layers}{Neurons}     & $10$  &     $50$    &   $100$   &   $150$  \\
    \midrule
    $1$ &   $3.73\%$    &   $2.50\%$      &     $1.73\%$      &   	$4.78\%$   \\
    $3$ &   $3.18\%$    &	$1.85\%$      &   	$3.61\%$      &   	$2.88\%$   \\
    $5$ &   $1.49\%$    &	$3.00\%$      &   	$2.81\%$      &     $3.25\%$   \\
    $7$ &   $2.28\%$    &	$2.83\%$      &     $3.80\%$      &     $2.70\%$   \\
    \bottomrule
  \end{tabular}
\end{table}

\begin{table}
 \caption{Hyper-parameters sweep with Fourier feature decoder, $p=150$.}\label{apdx_sweep_f150}
  \centering
  \begin{tabular}{c|cccc}
    \toprule
    \diagbox{Layers}{Neurons}     & $10$  &     $50$    &   $100$   &   $150$  \\
    \midrule
    $1$ &   $2.49\%$    &   $2.62\%$      &     $1.84\%$      &   	$2.58\%$   \\
    $3$ &   $2.19\%$    &	$1.64\%$      &   	$2.32\%$      &   	$1.63\%$   \\
    $5$ &   $1.93\%$    &	$2.67\%$      &   	$3.47\%$      &     $2.87\%$   \\
    $7$ &   $2.47\%$    &	$3.32\%$      &     $3.16\%$      &     $3.36\%$   \\
    \bottomrule
  \end{tabular}
\end{table}

\end{document}